\documentclass[journal,10pt]{IEEEtran}

\usepackage{graphicx}
\usepackage{subfigure}
\usepackage{booktabs} 
\usepackage{amsmath}
\usepackage{amssymb}
\usepackage{mathrsfs}
\usepackage{bm}
\usepackage{tablefootnote}
\usepackage{epigraph}
\usepackage{multirow}
\usepackage{bbding}
\usepackage{pifont}
\usepackage{soul}
\usepackage{cite}
\usepackage{color}
\usepackage[table]{xcolor}
\usepackage{url}
\usepackage{amsfonts}
\usepackage{wrapfig}
\usepackage{overpic}
\usepackage[pagebackref=false,breaklinks=true,colorlinks,bookmarks=false]{hyperref}

\usepackage{xspace}
\def\eg{\textit{e.g.}}
\def\ie{\textit{i.e.}}

\def\etal{\textit{et al. }}

\newcommand{\figref}[1]{Fig.~\ref{#1}}

\newcommand{\tabref}[1]{Tab.~\ref{#1}}

\newcommand{\nClass}{\ensuremath{K}}
\newcommand{\nStage}{\mathcal{Q}\xspace}

\newcommand{\mypartitletwo}[1]{\vspace*{-.5ex}~\\{\noindent \underline{\bf #1}}}
\newcommand{\mypartitle}[1]{\vspace*{-.5ex}~\\{\noindent \bf #1}}

\newcommand{\mapping}{\ensuremath{G}\xspace}

\newcommand{\params}{\ensuremath{\theta}\xspace}
\newcommand{\data}{\ensuremath{X}\xspace}

\newcommand{\featSpace}{\ensuremath{\mathrm{\cal X}}\xspace}
\newcommand{\lSpace}{\ensuremath{\mathrm{\cal Y}}\xspace}

\newcommand{\nsamples}{\ensuremath{N}\xspace}

\newcommand{\nTrees}{\ensuremath{M}\xspace}

\newcommand{\samp}{\ensuremath{\mathbf{x}}\xspace}

\newcommand{\func}{\mathcal{G}\xspace}
\newcommand{\lfunc}{\ensuremath{\mathbf{g}}\xspace}

\def\etal{{\em et al.~}}

\usepackage{silence}
\begin{document}
%
\title{Deep Negative Correlation Classification}
\author{Le Zhang, Qibin Hou, Yun Liu, Jia-Wang Bian, Xun Xu, Joey Tianyi Zhou and Ce Zhu,~\IEEEmembership{Fellow,~IEEE}
\thanks{Le Zhang and Ce Zhu are with the School of Information and Communication Engineering, University of Electronic Science and Technology of China.}
\thanks{Qibin Hou is with the School of Computer Science, Nankai University.}
\thanks{Jia-Wang Bian is with Department of Engineering Science, University of Oxford.}
\thanks{Yun Liu, Xun Xu and Joey Tianyi Zhou are with the Agency for Science, Technology and Research, Singapore.}
}





\markboth{Journal of \LaTeX\ Class Files,~Vol.~14, No.~8, August~2015}%
{Shell \MakeLowercase{\textit{et al.}}: Bare Demo of IEEEtran.cls for IEEE Transactions on Magnetics Journals}
%



\IEEEtitleabstractindextext{%
\begin{abstract}
Ensemble learning serves as a straightforward way to improve the performance of almost any machine learning algorithm. 
 Existing deep ensemble methods usually na\"ively train many different models and then aggregate their predictions. 
 This is not optimal in our view from two aspects: 
 i) Na\"ively training multiple models adds much more computational burden, especially in the deep learning era;
 ii) Purely optimizing each base model without considering their interactions limits the diversity of ensemble and  performance gains.
 We tackle these issues by proposing deep negative correlation classification (DNCC),
 in which the accuracy and diversity trade-off is systematically controlled by decomposing the loss function seamlessly into individual accuracy and the ``correlation'' between individual models and the ensemble. 
 DNCC yields a deep classification ensemble where the individual estimator is both accurate and ``negatively correlated''. 
 Thanks to the optimized diversities, DNCC works well even when utilizing a shared network backbone, which significantly improves its efficiency when compared with most existing ensemble systems. Extensive experiments on multiple benchmark datasets and network structures demonstrate the superiority of the proposed method. 
\end{abstract}

}

\maketitle

\IEEEdisplaynontitleabstractindextext

%
\IEEEpeerreviewmaketitle

\section{Introduction}\label{sec:introduction}
%
%
%
%
\IEEEPARstart{E}{nsemble} learning typically fuses multiple models to get better performance than its individual models.
It has been used in multiple research fields such as machine learning~\cite{breiman1996bagging,breiman2001random,rodriguez2006rotation,valentini2004bias},  computer vision~\cite{Zhang2020OrderlessReID,zhang2019nonlinear,avidan2007ensemble}, and so on.  
Dietterich~\cite{dietterich2000ensemble} explained
the success of ensemble learning from the statistical, computational, and
representational views. In addition,
bias-variance decomposition~\cite{zhang2019nonlinear,brown2005managing} and strength-correlation~\cite{breiman2001random} also shed light upon the rationale of ensemble learning theoretically.

Apart from the accuracy of the individual estimator, it is widely convinced that much of the success of ensemble learning is attributed to the degree of disagreement, or ``diversity'', within the system. A simple yet intuitive explanation is that millions of identical estimators are obviously no better than any individual amongst them. Under this umbrella, a frenzy of efforts has been devoted to encouraging better accuracy and diversity trade-offs. A typical solution is to fully optimize the individual estimator while injecting randomness into the ensemble, \eg, randomly manipulating the training data to provide each learner with a different subset of patterns or features~\cite{breiman2001random, rodriguez2006rotation}. 

\newcommand{\AddImg}[1]{%
\includegraphics[width=0.7\columnwidth, height=0.5\columnwidth]{#1}%
}

\begin{figure*}[!t]
    \centering
    \footnotesize
    \renewcommand{\arraystretch}{0.6}
    \setlength{\tabcolsep}{0.35mm}
    \begin{tabular}{cc}
          \AddImg{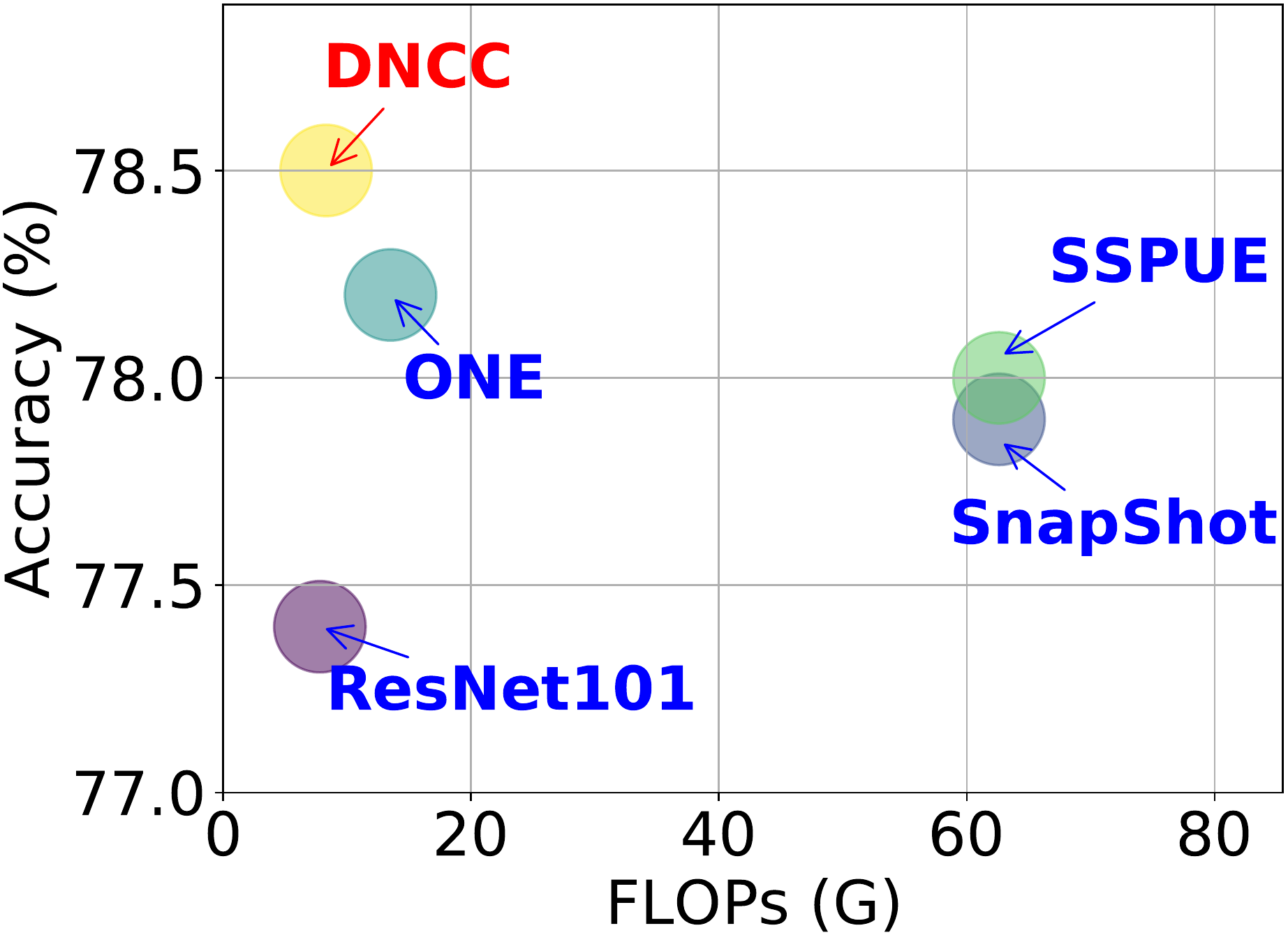} &
        \AddImg{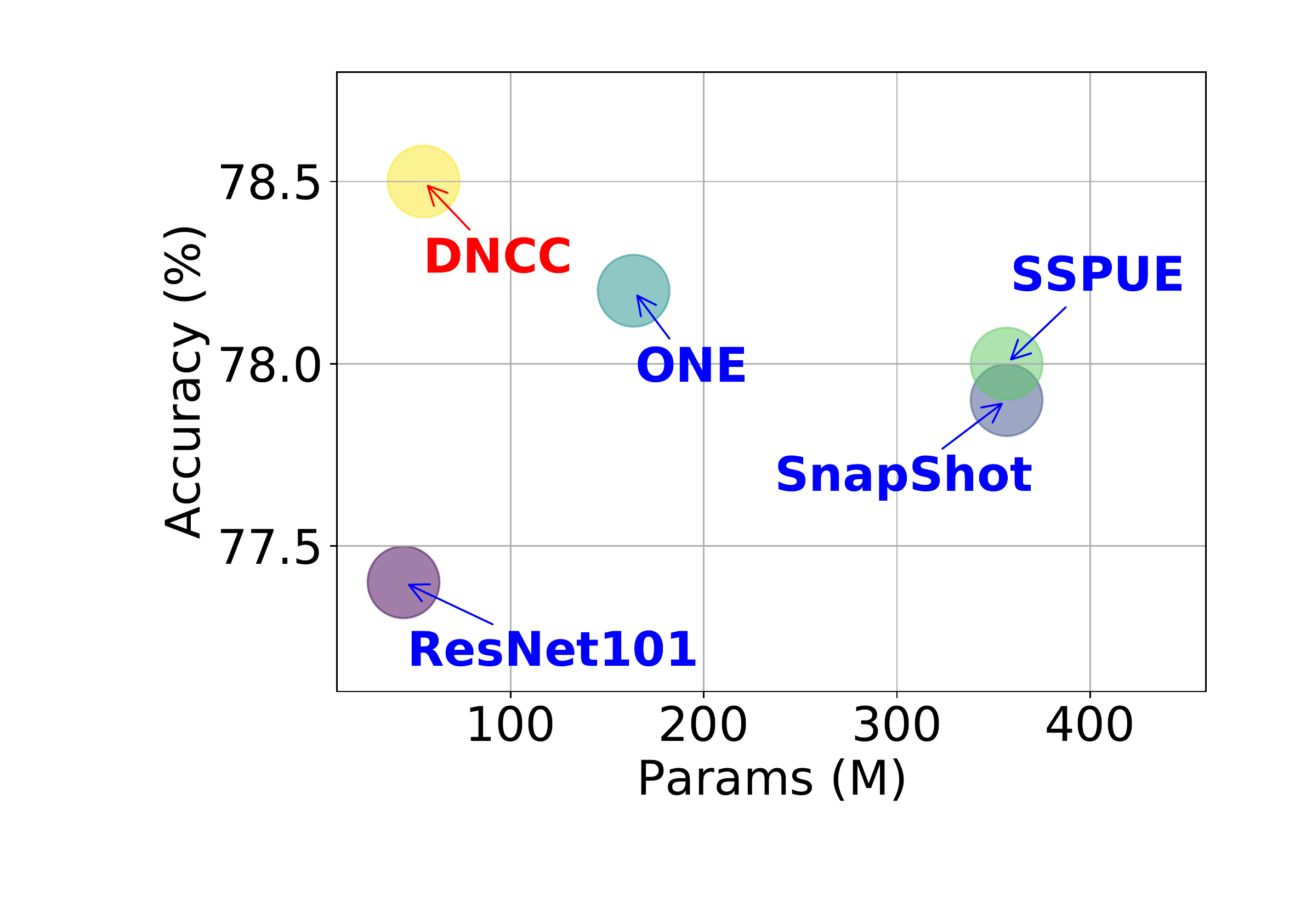} 
        \\
        (a)Accuracy-FLOPs trade-off. & (b) Accuracy-Params trade-off.
     \end{tabular}
    \vspace{2pt}
    \caption{Performance and network complexity trade-offs of different methods.  We choose ResNet101 as the baseline network and test them with the input size of $224\times224$. Obviously, DNCC obtains better trade-off between the performance and complexity. Please note that some methods utilize multiple networks only in either the training (\ie, ONE) or the inference stages (\ie, Snapshot). For fair comparisons, here we report the maximum number of parameters/FLOPs in both stages.}
    \label{fig:trade-off}
\end{figure*}%

However, the exact roles of accuracy and diversity in ensemble systems have not been well addressed although some theoretical analyses exist for the regression ensemble. Motivated by \textit{bias-variance} decomposition, \cite{ueda1996generalization} showed that the mean squared error (MSE) for ensemble can be further decomposed into the \textit{bias-variance-covariance} three-way trade-off. In this way, the optimal ensemble is said to be both ``accurate'' and ``diversified''. Based on the bias-variance-covariance, \cite{liu2000evolutionary} first proposed the well-known negative-correlation learning from the evolutionary computation point of view by explicitly managing the accuracy and diversity of the individual estimator.  This has motivated several works such as ~\cite{perales2021global,chen2018semisupervised}, and more recent ones in the deep learning era~\cite{zhang2019nonlinear,shi2018crowd,zhao2019enhancing}.

As for ensemble classification, where the individual estimators output discrete class labels, the ``diversity'' is not yet well understood and remains an open research issue. Although \cite{alhamdoosh2014fast} have made the first attempt by utilizing the one-hot coding on the category labels and training multiple models with the MSE under the negative correlation learning framework, this method is far from optimal. On one hand, the MSE is prone to outliers and is less robust than commonly used classification loss functions, \eg, Softmax Cross-Entropy loss in the deep learning era. On the other hand, this method has a high computational complexity owing to the computation of the pseudo-inverse of a large matrix. Therefore, existing ensemble classification methods mainly utilize heuristic strategies to enhance diversities implicitly. Examples include using different diversity measures~\cite{tang2006analysis}, randomly sampling data/feature subset~\cite{breiman2001random,rodriguez2006rotation}, utilizing different hyper-parameters~\cite{zhang2017benchmarking}, randomly dropping network activations/connections~\cite{hinton2012improving,wan2013regularization} and aggregating different network outputs along the optimization path~\cite{huang2017snapshot}. Bian~\etal provide some theoretical insights on the diversity measurement in~\cite{bian2021does}. However, their work is still lacking because it only focus on binary classification in shallow learning scenarios.

\begin{figure*}
    \centering
    \includegraphics[width=0.7\textwidth]{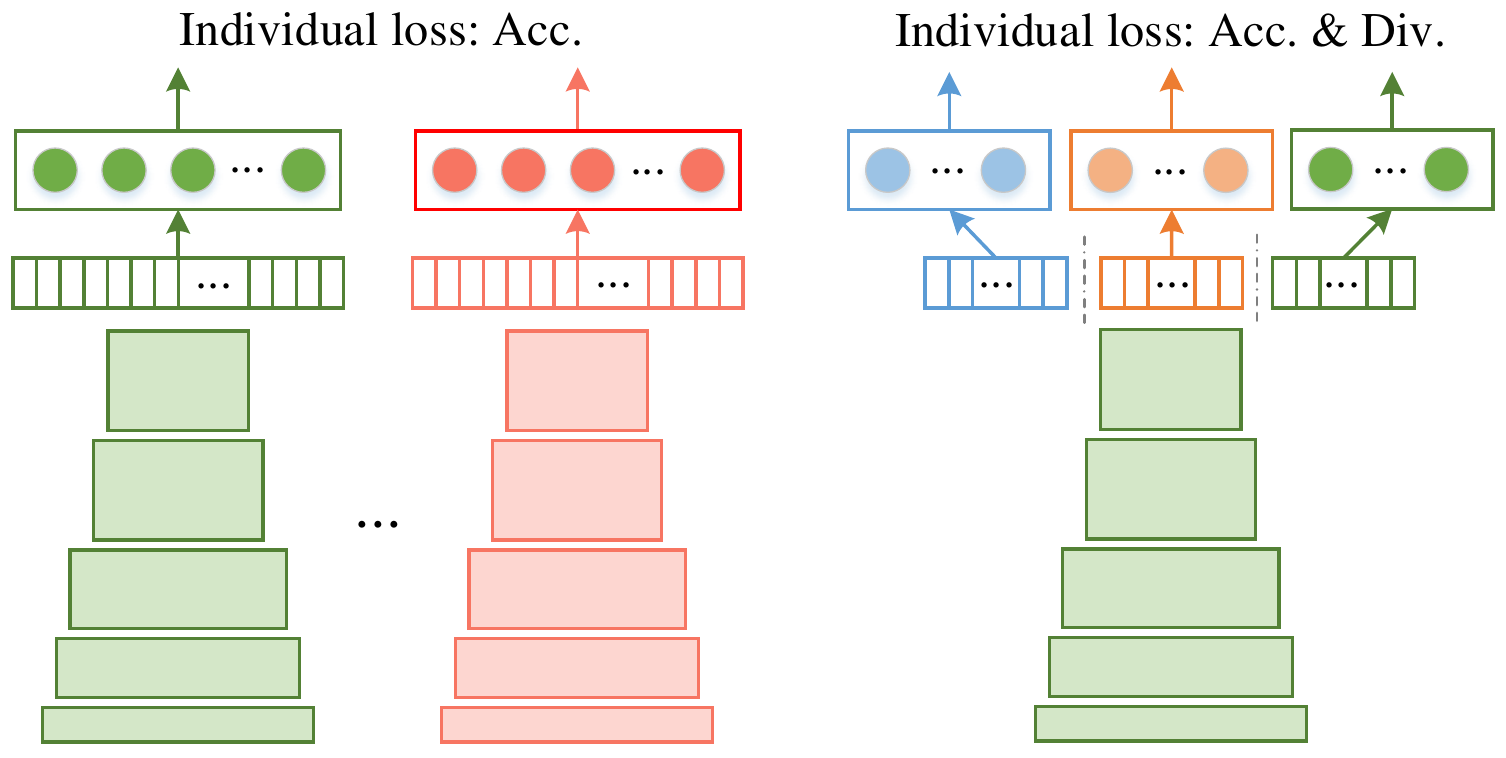}%
    \caption{Comparison between conventional ensemble methods and the proposed DNCC. Left: Conventional ensemble na\"ively trains many different models and optimizes individual accuracies independently. Right: DNCC jointly trains an efficient deep ensemble (with negligible extra parameters) where the individual estimator is both accurate and diversified.}
    \label{fig:comp1}
\end{figure*}

From the optimization point of view, a better way for ensemble classification is to jointly optimize the accuracy as well as the diversity explicitly and this has been barely studied for deep classification ensembles. In this work, we tap into this gap and propose deep negative correlation classification (DNCC) that is backbone-independent and end-to-end trainable for optimizing the long-standing accuracy-diversity trade-offs for classification ensemble. The main idea of DNCC, as illustrated in~\figref{fig:comp1}, is a new loss function for ensemble deep classification, inspired by the ``negative correlation learning''~\cite{liu2000evolutionary} which is commonly used in ensemble regression. More specifically, we seamlessly decompose the Softmax Cross-Entropy loss of ensemble deep networks into the individual loss of each network and their Bregman information, a quantity originally motivated by \textit{the rate-distortion theory} and used to measure the correlation amongst the ensemble here, and then derive a simple and efficient method for deep ensemble classification. The loss function is readily pluggable into any network architecture and amenable to training via backpropagation. Finally, we show that our DNCC outperforms challenging baselines on multiple benchmark datasets and network structures including CNNs \cite{he2016deep,huang2017densely}, Transformers \cite{liu2021swin}, and MLPs \cite{hou2021vision}.
\begin{itemize}
    \item We provide the definition of diversity in the deep classification ensembles by decomposing the commonly used soft-max crossentropy loss seamlessly into individual accuracy and the “correlations”. 
    \item Based on our framework, we show it is easy to optimize the accuracies and the the diversities of the base learners in an end-to-end manner. In this way,  the proposed method naturally yields both `` accurate" and ``diversified" deep ensembles.
    \item We demonstrate the effectiveness of our approach on different datasets when taking different network backbones. We show the proposed method is able to show consistent improvement over existing ensemble methods with significantly less FLOPs.
\end{itemize}
The rest of this paper is organized as follows. The related work is summarized in Section~\ref{headings}. And then, the definition of the diversity and the ``accuracy-diversity" decomposition is presented in Section~\ref{others}. Finally,
the empirical results are presented in Section~\ref{sec:exp}, followed by
the conclusion in Section~\ref{sec:conc}.

\section{Related Work} \label{headings}

\subsection{Conventional Ensemble Classification}
Representative conventional ensemble methods include bagging and boosting. They have been well studied in recent years and applied widely in different applications. \cite{breiman1996bagging} works by training multiple classifiers, which are formed by making bootstrap replicates of the learning set, using these as new learning sets, and then aggregating individual results. Due to the independence amongst the ensemble, each base model could be trained parallelly. As a special case of bagging, random forest~\cite{breiman2001random} utilizes multiple decision trees as the base classifier and demonstrates its superiority in a wide range of applications.  Boosting~\cite{freund1996experiments} works in a curriculum learning manner by first solving easy samples and progressively giving more focus to samples that are difficult to classify. Bian~\etal~\cite{bian2019ensemble} formulate ensemble pruning problem as an objection maximization problem
based on information entropy.  For more details on conventional ensemble classification, please refer to~\cite{ren2016ensemble}.

\subsection{Ensemble Deep Classification}
Although deep learning based methods have proven to surpass their shallow counterparts in various tasks, researchers have successfully shown that their performance could be further enhanced by ensemble learning. \cite{hinton2012improving} introduced a dropout strategy to prevent the co-adaptation of feature learners, in which the key idea is to randomly drop units (along with their connections) from a network during training. It can be seen as an extreme case of bagging and each parameter of the network is very strongly regularized by sharing it with the corresponding parameter in all the other models~\cite{hinton2012improving}. The adaptive version of dropout is proposed in~\cite{ba2013adaptive} where a binary belief network is overlaid on a network and is used to regularize its hidden units by selectively setting activities to zero. Motivated by dropout, \cite{wan2013regularization} introduced DropConnect to regularize large fully-connected layers within neural networks. It sets a randomly selected subset of weights within the network to zero and thus each unit essentially receives input from a random subset of units in the previous layer. DropConnect could be regarded as a larger ensemble of deep neural networks than dropout~\cite{wan2013regularization}. In~\cite{cirecsan2012multi}, multi-column structures are proposed where each column is actually a convolutional neural network (CNN) with different parameters, and outputs of all columns are averaged. 
The proposed method improves state-of-the-art performance on several benchmark datasets.  Moreover,  \cite{zhang2016visual} proposed an ensemble of randomized deep networks by the way of entropy minimization strategy~\cite{grandvalet2005semi} and achieved improved results in visual tracking. 

Stochastic multiple choice learning (sMCL) is proposed in~\cite{lee2016stochastic} to train diverse deep ensembles, which follows a ``winner-take-gradient'' training strategy. Experimental results demonstrate the broad applicability and efficacy of sMCL for training diverse deep ensembles. An ensemble of deep CNN is introduced in~\cite{Zhang2020OrderlessReID}, where the individual results are aggregated by the KemenyYoung method~\cite{levin1995introduction}.
Deep neural decision forest~\cite{kontschieder2015deep,rota2014neural} unifies random forest with the representation learning functionality from deep convolutional networks in an end-to-end manner. In \cite{zhou1702deep,pang2018improving}, the deep forest is proposed to generate a deep forest ensemble with a cascade structure that enables deep forest to do representation learning, and the number of cascade levels can be adaptively determined such that the model complexity can be automatically set. BatchEnsemble is established in~\cite{wenbatchensemble} where each weight matrix is defined as the Hadamard product of a shared weight among all ensemble members. Besides, \cite{shaham2016deep} showed how deep learning methods could be applied in the context of crowdsourcing and unsupervised ensemble learning. Ensemble-based Decorrelation Method is introduced in~\cite{gu2018regularizing} to regularize deep neural networks and avoid overfitting. Snapshot ensemble~\cite{huang2017snapshot} trains a single neural network converging to several local minima along its optimization path and saves the model parameters.  In~\cite{lakshminarayanan2017simple}, the authors introduced a simple and scalable predictive uncertainty estimation using Deep Ensembles. In the same way, other methods~\cite{gal2016dropout,ritter2018scalable,blundell2015weight} approximate Bayesian inference for neural networks with Bayesian model averaging. A naive on-the-fly-ensemble is introduced in~\cite{zhu2018knowledge}. It works by training a single multi-branch network while simultaneously establishing
a strong teacher on the fly by aggregating all the results to enhance the learning of the target network in a knowledge distilling strategy.  Random subspace strategy is used in ~\cite{zheng2018deep} for video classification. Chen~\etal~\cite{chen2021class} design a new loss function to rectify the bias toward the majority classes for class-imbalance deep learning.

Different from existing methods which usually implicitly encourage diversity in the ensemble system, we explicitly decompose the ensemble Softmax Cross-Entropy loss into individual classification loss and the pairwise correlations between individual predictions and the ensemble outputs. This is beneficial in the sense that both accuracy and diversity are fully optimized by back-propagation. To summarize, we make the following contributions:

\begin{figure*}
    \centering
    \begin{overpic}[width=\linewidth]{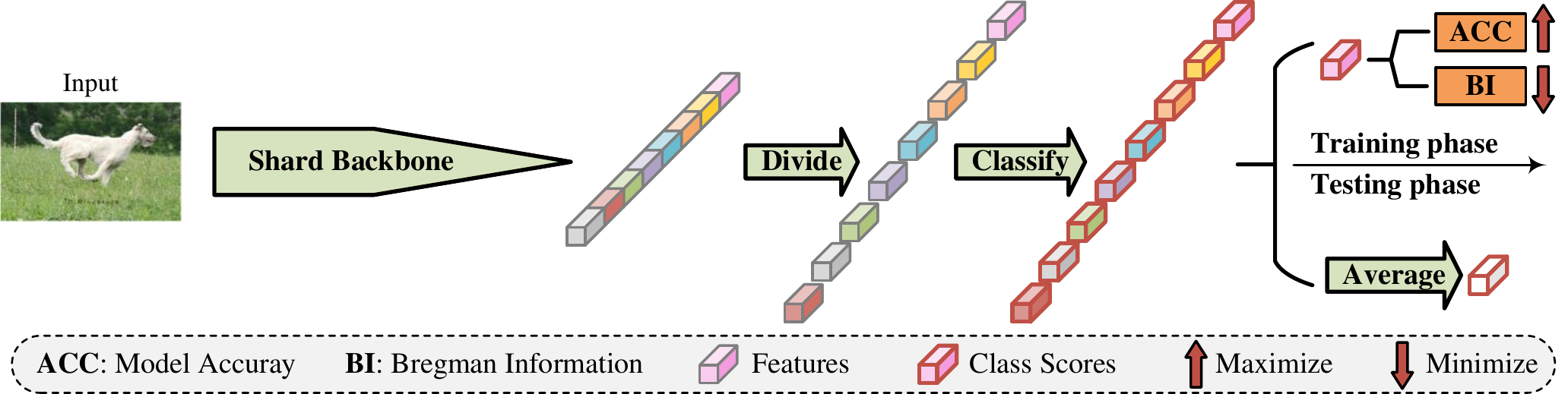}
        \put(45.2,21.7){$\func_{\nStage-1}$}
        \put(73.2,24.0){$\func_{\nStage}^m$}
        \put(99.5,20.5){$L_m$}
    \end{overpic}
    \caption{DNCC consists of multiple networks with a shared backbone. In the penultimate layer, we divide the features into multiple non-overlapping subsets and train multiple classifiers accordingly. In the training phase, each classifier is forced to be both accurate and ``negative correlated''. In the testing phase, we simply use the average aggregation.}
    \label{fig:dncc}
\end{figure*}

\section{Methodology}
\label{others}
Before elaborating on the proposed ensemble deep classification method,
we first briefly present the notations and background knowledge.
We assume that we have access to $\nsamples$ training samples, \ie,
$\data = \{\samp_1, \dots, \samp_\nsamples \}$. 
Our objective is to predict their category labels, \ie,
$Y= \{y_1,\dots, y_\nsamples \}$.
We denote a generic data point/feature tensor by $\samp$ and use $\samp_{\diamond}$,
with $\diamond$ denoting the place-holder for the index wherever necessary.
$y$ and $y_{\diamond}$ are similarly defined.
Suppose we have $\samp\in\featSpace$ and $y\in\lSpace=\{1,\cdots, \nClass\}$, in which $K$ is the number of classes.

We achieve our goal by learning a mapping function
$\mapping: \featSpace \rightarrow \lSpace$. Then the learning problem is to use the set $\data$ to learn a mapping function $\mapping$,
parameterized by $\params$, to approximate their label $Y$ as accurate as possible:
\begin{equation} \label{eq:int}
  L(\mapping)=-\frac{1}{\nsamples}\sum_{i=1}^{\nsamples}log(\frac{e^{f_{y_i}}}{\sum_{j=1}^{K}e^{f_j}}),
\end{equation}
where $f_j$ denotes the $j^{th}$ element ($j \in \{1,\cdots,\nClass\}$) of the vector of class scores $f$.  For simplicity, we will use $L$ to represent $L(G)$ whenever the dependence of the loss function with the parameters can be easily inferred from the context.
We consider the mapping function $\mapping$ to be an ensemble of deep networks, composed of 
 $\nTrees$ base classifier $\mathcal{G}= \{\func^m\}_{m=1}^{\nTrees}$,
where the classifiers $\func^m: \featSpace \rightarrow \lSpace, m \in \{1,\ldots, \nTrees\}$, called \emph{based deep networks}, are combined using averaging:
\begin{equation}
     \overline{L}=-\frac{1}{\nsamples}\sum_{i=1}^{\nsamples}log(\frac{1}{\nTrees}\sum_{m=1}^{\nTrees} \frac{e^{f^m_{y_i}}}{\sum_{j=1}^{K}e^{f^m_j}}),
     \label{ensemble}
 \end{equation}
 where $f^m_j$ denotes the $j^{th}$ element ($j \in \{1,\cdots,\nClass\}$) of the vector of the $m^{th}$ classifiers' class scores. Now we provide the definition of the Bregman divergence as follows:

\mypartitletwo{Definition 1.} (Bregman Divergence~\cite{bregman1967relaxation}). Let $\lfunc: \mathcal{S}\rightarrow \mathbb{R}, \mathcal{S}=dom(\lfunc) $ be a strictly
convex function defined on a convex set $\mathcal{S} \subseteq  \mathbb{R}^d$ such that $\lfunc$ is differentiable on $ri(\mathcal{S})$, assumed to
be nonempty. The \emph{Bregman divergence} $d_\func: \mathcal{S} \times ri(\mathcal{S}) \rightarrow [0,\infty)$ is defined as:
\begin{equation}
    d_\lfunc(t+\delta ,t)=\lfunc(t+\delta )-\lfunc(t)-<\delta ,\bigtriangledown\lfunc(t)>,
    \label{bregman}
\end{equation}
where $\bigtriangledown\lfunc(t)$ and $ ri(\mathcal{S}$) represent the gradient vector of $\lfunc$ evaluated at $t$ and the relative interior of $\mathcal{S}$, respectively. $dom(\lfunc)$ represents the effective
domain of $\lfunc$, \ie, set of all $t$ such that $|\lfunc(t)| < + \infty$ is denoted by $dom(\lfunc)$.

\mypartitletwo{Definition 2.} (Bregman Information~\cite{banerjee2005clustering}). Let $T$ be a random variable that takes values in $\mathcal{T} = \{t_i\}^\nsamples_{i=1}$ following a probability measure $\nu$. Let $\mu=E_\nu[T]=\sum_{i=1}^{\nsamples}\nu_it_i \in ri(\mathcal{S})$ and $d_\lfunc: \mathcal{S} \times ri(\mathcal{S}) \rightarrow [0,\infty)$ be the Bregman divergence.Then the \emph{Bregman information} of $T$ in terms of $d_\lfunc$ is defined as:
\begin{equation}
    I_\lfunc=E_\nu[d_\lfunc(T,\mu)]=\sum_{i=1}^{\nsamples}\nu_id_\lfunc(t_i,\mu).
    \label{information}
\end{equation}

With Definition 1 and 2, we have the following results:

\mypartitletwo{Lemma 1.} Given any convex function $\lfunc$, for any random variable $T$, we have:
\begin{equation}
    E[\lfunc(T)]-\lfunc(E[T])=I_\lfunc(T).
\end{equation}

\mypartitletwo{Proof.} 

\begin{equation}
\begin{aligned}
&E[\lfunc(T)]-\lfunc(E[T]]\\
&=E[\lfunc(T)]-\lfunc(E[T])-E[<T-E[T],\bigtriangledown\lfunc(E[T])>]\\
&=E[\lfunc(t)-\lfunc(E[T])-<T-E[T],\bigtriangledown\lfunc(E[T])>]\\
&=E[d_\lfunc(T,E[T])]\\
&=I_\lfunc(T)
\end{aligned}
\end{equation}


 With the above results, it is easy to obtain the following results by considering the convexity of the \emph{-log} function.
 
\mypartitletwo{Proposition 1.} For an ensemble of network, its Softmax Cross-entropy Loss $\overline{L}$, as defined in Eq.~\eqref{ensemble}, can be decomposed into the average loss of its base network and the Bregman Information:
\begin{equation}
    \overline{L}=\frac{1}{\nTrees}\sum_{m=1}^{\nTrees}L_m-I_{-log}.
    \label{decom}
\end{equation}


Proposition 1 explains the effect of error correlations in an ensemble system by stating that the Softmax Cross-entropy Loss of the ensemble network is guaranteed to be less than or equal to the average loss of the base networks. Existing ensemble classification methods mainly work by training multiple models independently. This may not be optimal because, as demonstrated in Proposition 1, the ensemble loss consists of both the individual loss and the non-negative Bregman information of the inputs. Based on this, we propose to learn a correlation-regularized ensemble system with the following objective:
\begin{equation}
\begin{aligned}
    L_m=&-\frac{1}{\nsamples}\sum_{i=1}^{\nsamples}[log({\frac{e^{f^m_{y_i}}}{\sum_{j=1}^{K}e^{f^m_j}}})\\
    &-\lambda*d_{-log} ({\frac{e^{f^m_{y_i}}}{\sum_{j=1}^{K}e^{f^m_j}}},\frac{1}{\nTrees}\sum_{m=1}^{\nTrees}{\frac{e^{f^m_{y_i}}}{\sum_{j=1}^{K}e^{f^m_j}}})],
    \label{loss}
    \end{aligned}
\end{equation}
where $d_{-log}$ could be obtained by setting $\lfunc$ as $-log$ in Definition 1. Eq.~\eqref{loss} can be regarded as a smoothed version of~Eq.~\eqref{decom} to improve the generalization ability of the ensemble models. 
The parameter $\lambda$ here controls the ensemble's accuracy and diversity and note that its optimal value may not necessarily be 1 because of the discrepancy between the training and testing data~\cite{brown2005managing,zhang2019nonlinear,zhao2019enhancing}.

More specifically, we consider the function $\func$ as an ensemble of networks as defined in~Eq.~\eqref{ensemble} where each base network is posed as:
\begin{equation}
\begin{aligned}
  \func^m(\samp_i)
  &=\func_{\nStage}^m(\mapping_{\nStage-1}^m\cdots(\func_1^m(\samp_i))),\\
  &m= 1,2\cdots\nTrees,~~i=1,2\cdots\nsamples,
\end{aligned}\label{func}
\end{equation}
where $m$, $i$, and $\nStage$ stand for the index for individual networks, 
the index for data samples and the depth of the network, respectively.
More specifically, each predictor in the ensemble consists of
cascades of feature extractors $\func_q^m$, $q=1,2\cdots\nStage-1$
and classifier $\func_{\nStage}^m$. 
As the diversity in the system are enhanced by regularizing the Bregman Information, we are able to use a shared network backbone for high efficiency. Formally, the lower levels of feature extractors are shared by each predictor,
\ie, $\func_q^m=\func_q$, $q=1,2\cdots,\nStage-1$, $m=1,2\cdots, \nTrees$.
 Based on that, we further divide the outputs of the highest level feature extractor $\func_{\nStage-1}$
to different subsets, each of which is used as input for different classifier $\func_{\nStage}^m$. This has been shown to be effective in generating an ensemble system without extra computational overhead than a standard single network~\cite{zhang2019nonlinear}. Apparently, this strategy is significantly more efficient than previous efforts in naively fusing multiple different deep networks~\cite{simonyan2014very,szegedy2015going}. An overview of the proposed method can be found in~\figref{fig:dncc} and illustrations on the accuracy and complexity trade-off are presented in \figref{fig:trade-off}.

\begin{table}[t]
\caption{Comparison of DNCC with other baseline methods on the CIFAR dataset.}
    \label{tab:cifar}
    \centering
    \small
    \begin{tabular}{l|c|c}
    \toprule[1pt]
         \textbf{Method}&\textbf{CIFAR10}&\textbf{CIFAR100} \\
         \midrule[0.5pt]
          Res50~\cite{he2016deep}&94.80&77.39\\
          Res50-Snapshot~\cite{huang2017snapshot}&94.78&78.47\\
          Res50-ONE~\cite{zhu2018knowledge}&94.89&78.56\\
          Res50-SSPUE~\cite{lakshminarayanan2017simple}&94.82&78.39\\
            Res50-DNCC&\textbf{95.05}&\textbf{79.05}\\
          \midrule[0.5pt]
          Res101~\cite{he2016deep}&94.98 &77.78\\
          Res101-Snapshot~\cite{huang2017snapshot}&95.30&78.52\\
          Res101-ONE~\cite{zhu2018knowledge}&95.41&78.61\\
          Res101-SSPUE~\cite{lakshminarayanan2017simple}&95.28&78.55\\
          Res101-DNCC&\textbf{95.53}&\textbf{78.82}\\
          \bottomrule[1pt]
    \end{tabular}
\end{table}

\subsection{Discussion}

Advocating both accuracy and diversities amongst individual models has been shown to be effective in the ensemble learning community. \cite{breiman2001random} derived the upper bound for ensemble's generation error by the way of both the ``strength'' and the ``correlation'' for base models. In addition, the Probably Approximately Correct (PAC) learning theory~\cite{valiant1984theory} shows that a good learner should be both accurate and with low hypothesis space complexity. In the proposed method, the complexity has been reduced in two ways. First, by using a shared network backbone, the proposed method is efficient and the complexity has been reduced significantly~\cite{zhang2019nonlinear}.  Second, the existing method~\cite{yu2011diversity} has shown that enhancing the ensemble diversity could also lead to a small hypothesis space complexity. The proposed loss function, as shown in~Eq.~\eqref{loss}, penalizes the pairwise correlation amongst the ensemble by reducing the Bregman distance between the individual networks and the ensemble results, and is thus beneficial in reducing the hypothesis space complexity.

\section{Experiments}\label{sec:exp}
To demonstrate the feasibility of DNCC, we evaluate it on several benchmark datasets, including CIFAR10~\cite{krizhevsky2009learning}, CIFAR100~\cite{krizhevsky2009learning}, and ImageNet~\cite{deng2009imagenet}. For CIFAR10 and CIFAR100, we employ the well-established residual networks (ResNet)~\cite{he2016deep} and use ResNet50 and ResNet100. We split the features of the last global average pooling into 8 non-overlapping subsets with equal dimensionality and train 8 classifiers accordingly. For CIFAR10, the batch size is set to 128 and we train the network for 150 epochs. The initial learning rate is 0.1 which is decreased by a factor of 0.1 for every 50 epochs. For CIFAR100, we train the network for 200 epochs and we decrease the learning rate by a factor of 0.1 at epochs 60, 120, and 160 with an initial learning rate of 0.1. 
As for the ImageNet dataset, apart from ResNet50 and ResNet101, we also evaluate DenseNet121~\cite{huang2017densely}. Moreover, we also add the recently proposed Transformer and MLP networks in the comparisons to further understand the merits of the proposed methods. For Transformer and MLP architectures, we choose the Swin-Transformer~\cite{liu2021swin} and the Vision Permutator~\cite{hou2021vision}. For all the networks trained on the ImageNet dataset, we firstly expand the features in the penultimate layer to $8\times$ of its original dimension, then split the resulting features into 8 non-overlapping subsets. Finally, we train 8 classifiers accordingly. For each network, we follow the original training protocol. For CNNs, we adaptively control the value of $\lambda$ by setting it to be $\frac{epoch}{\# epoch}\times 10^{-2}$, in which $epoch$ and $\#epoch$ stand for the number of current epoch and the number of the maximum epoch, respectively. For Swin-Transformers and Vision Permutator, $\lambda$ is simply set to be $5e-4$. We run all experiments with PyTorch~\cite{paszke2019pytorch}.

\subsection{Main Results}

We compared our proposed method with the baseline network (\ie Res50, Res101) and several state-of-the-art ensemble learning strategies. Firstly, we consider the Snapshot Ensemble~\cite{huang2017snapshot}. It trains a single neural network and saves the model parameters when converging to several local minima along the optimization path. We also consider the On-the-Fly Native Ensemble (ONE)~\cite{zhu2018knowledge}.  ONE trains multi-branch networks and distills the knowledge from the ensemble results to each branch on the fly. The other method we compare is the SSPUE (Simple and Scalable Predictive Uncertainty
Estimation using Deep Ensembles) which is a combination
of ensembles and adversarial training~\cite{lakshminarayanan2017simple}. 

\mypartitle{CIFAR10}.
The CIFAR10 dataset~\cite{krizhevsky2009learning} consists of natural color images sized at $32\times32$ pixels. 
It has 50,000 training images and 10,000 testing images from 10 classes. 
We use a standard data augmentation scheme~\cite{lin2013network}, in which the images are zero-padded with 4 pixels on each side and then randomly cropped to generate $32\times32$ images. 
Besides, we also horizontally flip the inputs with a probability of 0.5. 
We use ResNet50 and ResNet101~\cite{he2016deep}.
The experimental results are summarized in \tabref{tab:cifar}. From the experimental results, we can see that with the same network backbone, the proposed DNCC method yields the best results for both ResNet50 and ResNet101 and surpasses state-of-the-art ensemble learning strategies such as Snapshot Ensemble~\cite{huang2017snapshot}, ONE~\cite{zhu2018knowledge} and SSPUE~\cite{lakshminarayanan2017simple}.






\mypartitle{CIFAR100}. The CIFAR100 dataset~\cite{krizhevsky2009learning} has the same statistics as CIFAR10 except that the images are sampled from 100 classes. 
We use the same data augmentation technique as done for CIFAR10. 
In the same way, we use ResNet50 and ResNet101~\cite{he2016deep}. Results for Snapshot Ensemble, ONE, and SSPUE with the same architecture are also provided. Results are presented in \tabref{tab:cifar}. 
Conclusions from the CIFAR10 dataset are also applicable here.

\mypartitle{ImageNet}. 
The ILSVRC 2012 classification dataset~\cite{deng2009imagenet} consists of 1000 images classes, with a total of 1.2 million training images and 50,000 validation images. 
We adopt the same data augmentation scheme as in \cite{he2016deep} and apply a $224\times224$ center crop to images for testing. For this dataset, we use various architectures including CNN, Transformer, and MLP. As for CNN,we use ResNet50, ResNet101, and DenseNet121. Results for Snapshot Ensemble, ONE, and SSPUE with the same architecture for each network are also provided Results are summarized in \tabref{tab:Results}. In order to further understand the merits of DNCC, we also report the number of parameters and FLOPs (with input size of $224\times224$) of each method. Please note that some methods, such as ONE, use multiple auxiliary heads in the training phase and prune them in the inference phase. In addition, other methods, such as the Snapshot Ensemble, save multiple models and use all of them in the inference phase. For more fair comparisons, we report the maximum number of parameters and FLOPs for each method in both training and testing stages. It is also straightforward to see that DNCC
yield better trade-offs between the performances and network complexities.
\\
In \tabref{tab:Results-non-cnn}, we also evaluate DNCC on more advanced non-CNN architectures. We choose the Swin-Transformer \cite{liu2021swin} and the Vision Permutator (ViP)~\cite{hou2021vision} as the representative work of the transformer and MLP respectively. As the Swin-Transformer and the Vision Permutator typically need more training epochs (\ie, 300) to converge, we did not compare other ensemble methods because they typically need more FLOPs/Parameters and thus significantly slow down the training process. 
We can show that the proposed DNCC improves different baselines.  

\begin{table}[tp]
	\begin{center}
		\caption{
		Comparison of DNCC with other baseline methods on the ImageNet dataset.
		}
		\small
		\setlength{\tabcolsep}{8.3pt}
		\label{tab:Results}
		\begin{tabular}{l|c|c|c}
			\toprule[1pt]
			\textbf{Network} & \textbf{Accuracy} & \textbf{Params} & \textbf{FLOPs}\\ \midrule[0.5pt]
	Res50~\cite{he2016deep} & 76.1&\textbf{25.6}M&\textbf{4.1G}\\
  Res50-Snapshot~&76.4&204.8M&32.8G\\
          Res50-ONE&76.6&144.7&9.8G\\
          Res50-SSPUE&76.3&204.8M&32.8G\\
          Res50-DNCC&\textbf{76.8}&36.0M&4.5G\\
  
  \midrule[0.5pt]
  Res101~\cite{he2016deep} & 77.4&\textbf{44.6}M&\textbf{7.8G}\\
  Res101-Snapshot&77.9&356.8M&62.6G\\
          Res101-ONE&78.2&163.7M&13.5G\\
          Res101-SSPUE&78.0&356.8M&62.6G\\
           Res101-DNCC&\textbf{78.5}&55.0M&8.3G\\
  \midrule[0.5pt]
  	Dense121~\cite{huang2017densely} & 74.4&\textbf{8.0M}&\textbf{2.9G}\\
  Dense121-Snapshot&74.7&63.8M&23.0G\\
  Dense121-ONE&75.0&26.8M&4.4G\\
  Dense121-SSPUE&74.8&63.8M&23.0G\\
   Dense121-DNCC&\textbf{75.2}&23.6M&\textbf{2.9G}\\
  \bottomrule[1pt]
		\end{tabular}
	\end{center}
\end{table}

\begin{table}[!thb]
	\begin{center}
		\caption{
		Effect of DNCC on popular transformer and MLP-like models on the ImageNet dataset.
		}
		\small
		\setlength{\tabcolsep}{8.3pt}
		\label{tab:Results-non-cnn}
		\begin{tabular}{l|c|c|c}
			\toprule[1pt]
			\textbf{Network} & \textbf{Accuracy} & \textbf{Params} & \textbf{FLOPs}\\ \midrule[0.5pt]
   Swin-T~\cite{liu2021swin}&81.2&\textbf{28.3}M&\textbf{4.5G}\\
  Swin-T-DNCC&\textbf{81.5}&38.4M&\textbf{4.5G}\\
  \midrule[0.5pt]
   ViP-Small/7~\cite{hou2021vision} &81.5&\textbf{25.1M}&\textbf{6.9G}\\
   ViP-Small/7-DNCC &\textbf{81.9}&29.4M&\textbf{6.9G}\\
  \bottomrule[1pt]
		\end{tabular}
	\end{center}
\end{table}

\subsection{Trade-off between Accuracy and Diversity} 
In order to further understand the merits of DNCC, we shed light upon the trade-off between accuracy and diversity in this section.
In~\cite{diez2015diversity}, the authors showed that enhancing diversity could in principle lead to a small hypothesis space complexity which is essential in improving the generalization ability of the learning system. 
Here we compare the accuracy and diversities of DNCC with the conventional ensemble. 
In this study, we compare Res101-DNCC and Res101-Ensemble, which is trained by setting $\lambda=0$, and train them on the CIFAR10 dataset. 
We split the training data into two non-overlapping subsets with a ratio of 4:1 and use them to train and validate the methods, respectively. 
Following the previous section, the ensemble size is set to 8. 
Motivated by~\cite{rodriguez2006rotation}, we compare the pairwise accuracy and diversity of both methods. 
For the $i^{th}$ and $j^{th}$ classifier in the ensemble, the accuracy is measured by the mean accuracy of both classifiers, and the diversity is defined as:
\begin{equation}
  Diversity(i,j)= \frac{1}{K}\sum_{k=1}^{K}(1-\frac{W^i_k*W^j_k}{||W^i_k||*||W^j_k||}),
  \label{pair-diversity}
  \end{equation}
where $W^i_k$ is the weights of the $i^{th}$ classifier for the $k^{th}$ class.
Eq.~\eqref{pair-diversity} is essentially the average angle between the corresponding decision hyper-planes for the two classifiers~\cite{diez2015diversity}.

\begin{figure*}[!t]
    \centering
    \footnotesize
    \renewcommand{\arraystretch}{0.6}
    \setlength{\tabcolsep}{0.35mm}
    \begin{tabular}{cc}
        \AddImg{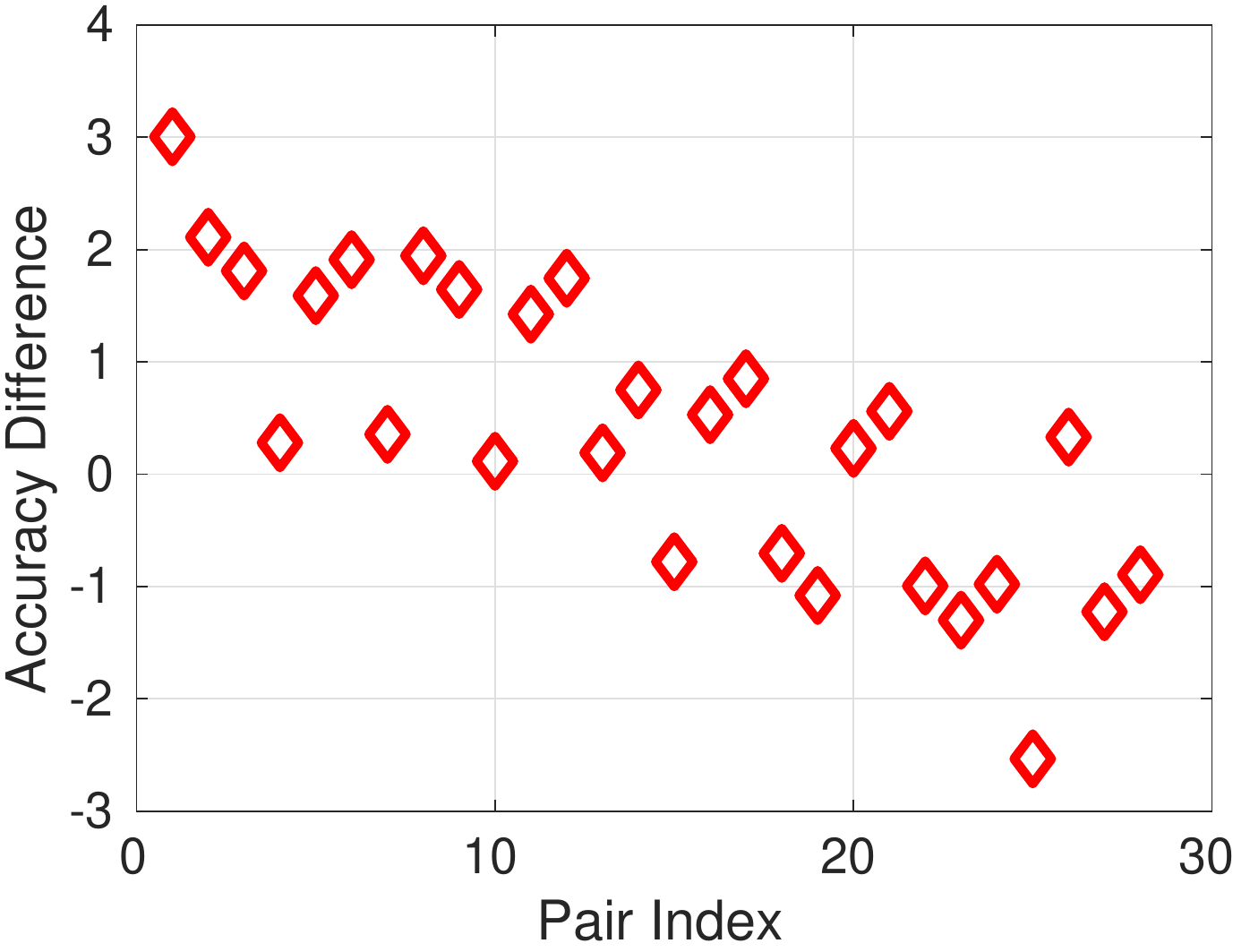} &
        \AddImg{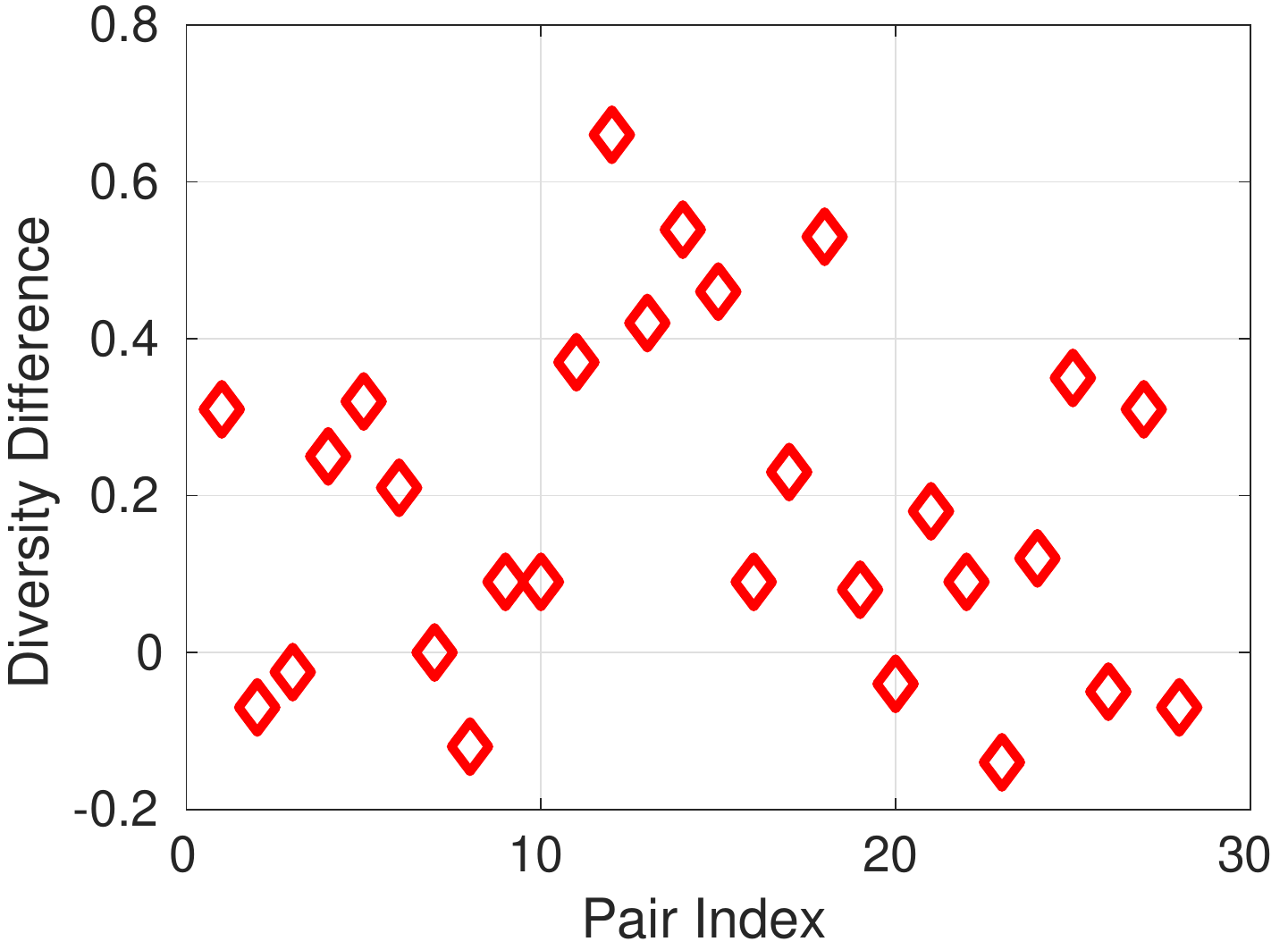} 
        \\
        (a) Accuracy improvement. & (b) Diversity improvement.
     \end{tabular}
    \vspace{2pt}
    \caption{Diversity and accuracy comparison between DNCC and the conventional ensemble on the CIFAR10 dataset. The accuracy is calculated by the mean accuracy of each pair and the diversity is defined in Eq.~\eqref{pair-diversity}. For figures (a) and (b), a positive value in the $y$-axis indicates that DNCC has better pairwise accuracy and diversity over the conventional ensemble, respectively. The experimental results indicate that, for each pair of classifiers, DNCC achieves competitive accuracy yet much better diversities over the conventional ensemble.}
    \label{fig:diversity-accuracy}
\end{figure*}

In this example, as there are 8 individual classifiers in the ensemble, there exist $C_8^2=28$ pairs and we visualize the diversity and accuracy for DNCC and the conventional ensemble in~\figref{fig:diversity-accuracy}. 
More specifically, in~\figref{fig:diversity-accuracy}(a) and~\figref{fig:diversity-accuracy}(b), we visualize the diversity and accuracy trade-off of both methods. 
The $x$-axis and $y$-axis stand for the accuracy and diversity improvement (positive value in the $y$-axis indicates that DNCC has better pairwise accuracy and diversity over the conventional ensemble, respectively) for each pair, respectively. \\
For the diversity part, DNCC shows clear superiority over the conventional ensemble baseline, as expected.
~\figref{fig:diversity-accuracy}(a) compares the accuracy improvement of DNCC over the conventional ensemble. 
The $x$-axis stands for the index of each pair. The $y$-axis is the accuracy difference between DNCC and the conventional ensemble. 
In this case, a positive value indicates that DNCC achieves better pairwise mean accuracy than the conventional ensemble. 
It is interesting that in most cases, the pairwise mean accuracy of DNCC is better than the conventional ensemble. 
Two reasons could lead to this improvement: i) The penalty term introduced in Eq.~\eqref{loss} could work as a strong regularizer for the individual models in the ensemble and thus reduce its over-fitting. ii) Advocating the diversity of DNCC in our setting could also lead to more diversified feature representation in the bottom network backbone, which essentially prevents their output units from co-adapting.
Similarly, \figref{fig:diversity-accuracy}(b) shows the diversity difference in the same manner.
Obviously, DNCC demonstrates a clear advantage in the diversity part over the conventional ensemble.
Finally, we observe that the DNCC and the conventional ensemble achieve $94.26\%$ and $93.97\%$ accuracy, respectively. 
However, for the individual classifier in both methods, the accuracy is mostly within $60\%-90\%$, which again demonstrates the effectiveness of ensemble methods in improving the final performance.

\subsection{Trade-off between Efficiency and Accuracy }
Previous ensemble methods~\cite{simonyan2014very,szegedy2015going} fuse outputs from $\nTrees$ different models in the inference phase,
and hence requiring $\nTrees$ times of computational overhead than a standard single model.
In contrast, the proposed method is able to have the same inference time in our design~\cite{zhang2019nonlinear}.
The speed of a single model using different backbones can be found in~\cite{tan2019efficientnet}. 

Although we mainly advocate an efficient solution of ``train 1 and get $N$ for free'', in practice, one could simply further enhance the classification accuracy by having the base network branch at an earlier stage. To demonstrated this, we have done extra experiments on the CIFAR100 dataset with the ResNet50 backbone. In particular, we grew different network branches after the Res2X, Res3X, Res4X blocks. We achieved the accuracy of 80.04\%, 79.70\%, and 79.05\%, respectively. In addition, the accuracy of 80.26\% was obtained in a na\"ive setting in which 8 independent ResNet50 were used. Those results are also provided in~\tabref{tab:share}. The results verify that inserting more diversities in the ensemble, by having different randomly initialized branches at the early stage of the network, could lead to better ensemble performances. 
\begin{table}[!thb]
	\begin{center}
		\caption{
		Accuracy-Efficiency Trade-off of DNCC  on CiFAR100 with a ResNet50 backbone.
		}
		\small
		\setlength{\tabcolsep}{8.3pt}
	\label{tab:share}
		\begin{tabular}{l|c|c}
			\toprule
			\textbf{Split Position} & \textbf{Shared Params}  &\textbf{ Accuracy} \\ \midrule
  Input&0$\%$&\textbf{80.26}$\%$\\
   Res2X&6$\%$&80.04$\%$\\
    Res3X&36$\%$&79.70$\%$\\
     Res4X&91$\%$&79.05$\%$\\
  \bottomrule
		\end{tabular}
	\end{center}
\end{table}

\subsection{Ablation Studies}
In this section, we provide some ablation studies to further understand the merits of the proposed DNCC. In particular, we study the effect of the ensemble size $N$ and the regularization parameter $\lambda$ in the following sections.

\mypartitle{Effect of the Ensemble Size}:
The existing ensemble learning theory~\cite{breiman2001random} shows that it will bring no harm in terms of the final accuracy to increase the ensemble size. However, this improvement does not come with no cost: increasing the ensemble size will dramatically increase the computational complexity as well. Therefore, in practice, one may need to control the ensemble size to achieve a better trade-off between performance and computational resources.

\begin{figure*}[!t]
    \centering
    \footnotesize
    \renewcommand{\arraystretch}{0.6}
    \setlength{\tabcolsep}{0.35mm}
    \begin{tabular}{cc}
        \AddImg{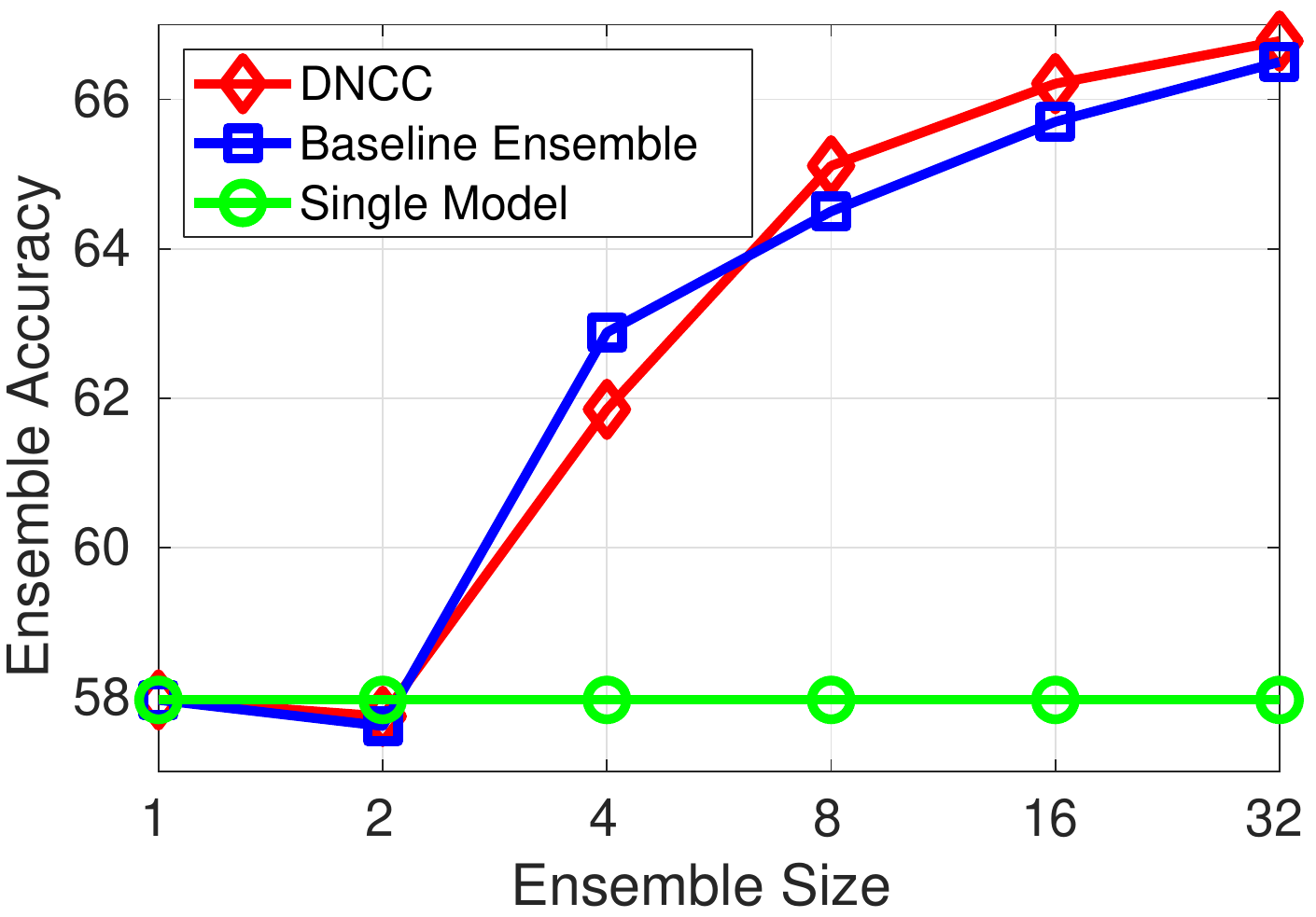} &
        \AddImg{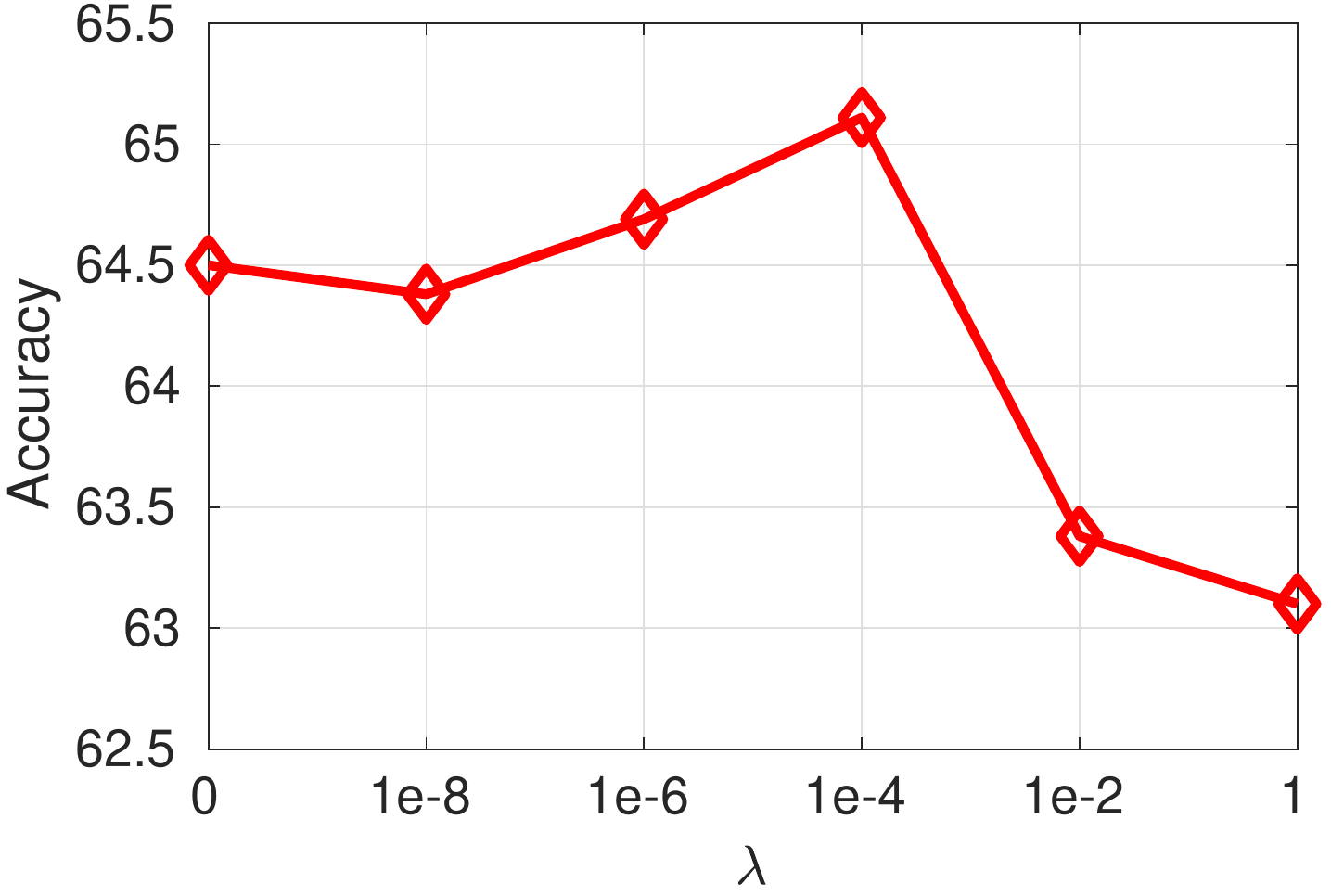}\label{fig:lambda}
        \\
        (a) Effect of the ensemble size. & (b) Effect of the parameter $\lambda$.
     \end{tabular}
    \vspace{2pt}
    \caption{Effect of the ensemble size $N$ and the parameter $\lambda$ on the CIFAR100 dataset. The figure (a) shows that both DNCC and the conventional ensemble perform better when a larger ensemble size is utilized. Furthermore, with a larger ensemble size ($\geq 4$), DNCC outperforms the conventional ensemble. The figure (b) also indicates that setting $\lambda$ to a small value of 1$e$-4 could yield improved results.}
    \label{fig:ensemble-lambda}
\end{figure*}%


In order to investigate the role of the ensemble size in DNCC, we conduct a set of experiments on the CIFAR100 dataset. 
We use a tight version of ResNet18 \cite{he2016deep} which we call \emph{tight-ResNet18} for this experiment. 
More specifically, for each residual block of ResNet18, we reduce the number of output channels by a factor of 8 to save the training time. 
We train $\nTrees$ different networks with the same training protocol with different weight initialization. 
All the networks are trained for 300 epochs with an initial learning rate of 0.1 and a batch size of 128. 
We decrease the learning rate by a factor of 0.1 at epochs 60, 120, and 160, respectively. 
$\lambda$ in Eq.~\eqref{loss} is set to 1$e$-4. 
\figref{fig:ensemble-lambda}(a) displays the performance of \emph{tight-ResNet18} ensemble as the effective ensemble size, $\nTrees$, is varied. 
It can be seen that both methods yield better performance than the single model in most cases and most importantly, DNCC performs better than the conventional ensemble method when we have $\nsamples>4$, which demonstrates the effectiveness of the proposed DNCC.

\mypartitle{Effective of $\lambda$}: 
The parameter $\lambda$ controls the correlation among base models in the ensemble system. 
On the one hand, setting $\lambda=0$ is equivalent to training each classifier in an independent manner.
On the other hand, a larger value of $\lambda$ could yield a less-correlated ensemble with high diversities. 
However, we also observe that a larger value of $\lambda$ could also have a negative effect on the accuracy of individual models, which could lead to worse final accuracy. 
This is because a larger value of $\lambda$ will affect the network optimization towards maximizing accuracy by weakening the effect of the first term in Eq.~\eqref{loss}.
In order to study the effect of $\lambda$, we conduct ablation experiments with different values of $\lambda$ using the previous settings. 
More specifically, we train 8 \emph{tight-ResNet18} models with the parameter $\lambda$ in $\{0,1e\text{-8},1e\text{-6},1e\text{-4},1e\text{-2},1 \}$ and report the results in \figref{fig:ensemble-lambda}(b). 
As can be observed, setting $\lambda$ to a small value of 1$e$-4 could yield the best result. In addition, we also report the results of different backbones on different datasets when setting $\lambda=0$ in \tabref{tab:eff:lam}. Results clearly show that the proposed DNCC outperforms the conventional ensemble by setting $\lambda=0$. Hence, it is beneficial to manage the accuracy-diversity amongst the ensemble. 

\begin{table}[t]
\caption{Comparison of DNCC with the conventional ensemble by setting $\lambda=0$.}
    \label{tab:eff:lam}
    \centering
    \small
    \begin{tabular}{l|c|c|c}
    \toprule[1pt]
         \textbf{Method}&\textbf{CIFAR10}&\textbf{CIFAR100}&\textbf{ImageNet} \\
         \midrule[0.5pt]
          Res50~\cite{he2016deep}&94.80&77.39&76.14\\
           Res50-Ensemble&94.85&78.42&76.51\\
           Res50-DNCC&\textbf{95.05}&\textbf{79.05}&\textbf{76.82}\\
          \midrule[0.5pt]
          Res101~\cite{he2016deep}&94.98&77.78&77.37\\
          Res101-Ensemble&95.36&78.48&78.12\\
          Res101-DNCC&\textbf{95.53}&\textbf{78.82}&\textbf{78.45}\\
          \bottomrule[1pt]
    \end{tabular}
\end{table}
\section{Conclusion}\label{sec:conc}
In this paper, we have presented the deep negative correlation classification (DNCC) algorithm to learn efficient and structure-independent deep network ensembles by involving Bregman information. 
Our analysis presents a new view of ensemble Softmax Cross-Entropy loss by decomposing it into individual accuracy and diversity between individual prediction and the ensemble output. 
Apart from the high efficiency, our proposed method is also advantageous when compared with existing ensemble methods by jointly optimizing the accuracy and diversities through back-propagation. 
Extensive experiments have shown the feasibility of the proposed method with multiple network structures on multiple benchmark datasets. 
Our work would be valuable in developing new accurate yet efficient deep ensemble learning algorithms. One limitation of our work is, for more complicated classification problems such as ImageNet, it may need an extra, but still affordable, amount of parameters.


%
\bibliography{reference}

\begin{thebibliography}{10}
\providecommand{\url}[1]{#1}
\csname url@samestyle\endcsname
\providecommand{\newblock}{\relax}
\providecommand{\bibinfo}[2]{#2}
\providecommand{\BIBentrySTDinterwordspacing}{\spaceskip=0pt\relax}
\providecommand{\BIBentryALTinterwordstretchfactor}{4}
\providecommand{\BIBentryALTinterwordspacing}{\spaceskip=\fontdimen2\font plus
\BIBentryALTinterwordstretchfactor\fontdimen3\font minus
  \fontdimen4\font\relax}
\providecommand{\BIBforeignlanguage}[2]{{%
\expandafter\ifx\csname l@#1\endcsname\relax
\typeout{** WARNING: IEEEtran.bst: No hyphenation pattern has been}%
\typeout{** loaded for the language `#1'. Using the pattern for}%
\typeout{** the default language instead.}%
\else
\language=\csname l@#1\endcsname
\fi
#2}}
\providecommand{\BIBdecl}{\relax}
\BIBdecl

\bibitem{breiman1996bagging}
L.~Breiman, ``Bagging predictors,'' \emph{Machine Learning}, vol.~24, no.~2,
  pp. 123--140, 1996.

\bibitem{breiman2001random}
------, ``Random forests,'' \emph{Machine Learning}, vol.~45, no.~1, pp. 5--32,
  2001.

\bibitem{rodriguez2006rotation}
J.~J. Rodriguez, L.~I. Kuncheva, and C.~J. Alonso, ``Rotation forest: A new
  classifier ensemble method,'' \emph{IEEE Transactions on Pattern Analysis and
  Machine Intelligence}, vol.~28, no.~10, pp. 1619--1630, 2006.

\bibitem{valentini2004bias}
G.~Valentini and T.~G. Dietterich, ``Bias-variance analysis of support vector
  machines for the development of {SVM}-based ensemble methods,'' \emph{Journal
  of Machine Learning Research}, vol.~5, no. Jul, pp. 725--775, 2004.

\bibitem{Zhang2020OrderlessReID}
L.~Zhang, Z.~Shi, J.~T. Zhou, M.-M. Cheng, Y.~Liu, J.-W. Bian, Z.~Zeng, and
  C.~Shen, ``Ordered or orderless: A revisit for video based person
  re-identification,'' \emph{IEEE Transactions on Pattern Analysis and Machine
  Intelligence}, 2020.

\bibitem{zhang2019nonlinear}
L.~Zhang, Z.~Shi, M.-M. Cheng, Y.~Liu, J.-W. Bian, J.~T. Zhou, G.~Zheng, and
  Z.~Zeng, ``Nonlinear regression via deep negative correlation learning,''
  \emph{IEEE Transactions on Pattern Analysis and Machine Intelligence}, 2019.

\bibitem{avidan2007ensemble}
S.~Avidan, ``Ensemble tracking,'' \emph{IEEE Transactions on Pattern Analysis
  and Machine Intelligence}, vol.~29, no.~2, pp. 261--271, 2007.

\bibitem{dietterich2000ensemble}
T.~G. Dietterich, ``Ensemble methods in machine learning,'' in
  \emph{International Workshop on Multiple Classifier Systems}.\hskip 1em plus
  0.5em minus 0.4em\relax Springer, 2000, pp. 1--15.

\bibitem{brown2005managing}
G.~Brown, J.~L. Wyatt, and P.~Ti{\v{n}}o, ``Managing diversity in regression
  ensembles,'' \emph{Journal of Machine Learning Research}, vol.~6, no. Sep,
  pp. 1621--1650, 2005.

\bibitem{ueda1996generalization}
N.~Ueda and R.~Nakano, ``Generalization error of ensemble estimators,'' in
  \emph{International Conference on Neural Networks}, vol.~1.\hskip 1em plus
  0.5em minus 0.4em\relax IEEE, 1996, pp. 90--95.

\bibitem{liu2000evolutionary}
Y.~Liu, X.~Yao, and T.~Higuchi, ``Evolutionary ensembles with negative
  correlation learning,'' \emph{IEEE Transactions on Evolutionary Computation},
  vol.~4, no.~4, pp. 380--387, 2000.

\bibitem{perales2021global}
C.~Perales-Gonz{\'a}lez, F.~Fern{\'a}ndez-Navarro, M.~Carbonero-Ruz, and
  J.~P{\'e}rez-Rodr{\'\i}guez, ``Global negative correlation learning: A
  unified framework for global optimization of ensemble models,'' \emph{IEEE
  Transactions on Neural Networks and Learning Systems}, 2021.

\bibitem{chen2018semisupervised}
H.~Chen, B.~Jiang, and X.~Yao, ``Semisupervised negative correlation
  learning,'' \emph{IEEE Transactions on Neural Networks and Learning Systems},
  vol.~29, no.~11, pp. 5366--5379, 2018.

\bibitem{shi2018crowd}
Z.~Shi, L.~Zhang, Y.~Liu, X.~Cao, Y.~Ye, M.-M. Cheng, and G.~Zheng, ``Crowd
  counting with deep negative correlation learning,'' in \emph{IEEE Conference
  on Computer Vision and Pattern Recognition}, 2018, pp. 5382--5390.

\bibitem{zhao2019enhancing}
W.~Zhao, B.~Zheng, Q.~Lin, and H.~Lu, ``Enhancing diversity of defocus blur
  detectors via cross-ensemble network,'' in \emph{IEEE Conference on Computer
  Vision and Pattern Recognition}, 2019, pp. 8905--8913.

\bibitem{alhamdoosh2014fast}
M.~Alhamdoosh and D.~Wang, ``Fast decorrelated neural network ensembles with
  random weights,'' \emph{Information Sciences}, vol. 264, pp. 104--117, 2014.

\bibitem{tang2006analysis}
E.~K. Tang, P.~N. Suganthan, and X.~Yao, ``An analysis of diversity measures,''
  \emph{Machine Learning}, vol.~65, no.~1, pp. 247--271, 2006.

\bibitem{zhang2017benchmarking}
L.~Zhang and P.~N. Suganthan, ``Benchmarking ensemble classifiers with novel
  co-trained kernel ridge regression and random vector functional link
  ensembles [research frontier],'' \emph{IEEE Computational Intelligence
  Magazine}, vol.~12, no.~4, pp. 61--72, 2017.

\bibitem{hinton2012improving}
G.~E. Hinton, N.~Srivastava, A.~Krizhevsky, I.~Sutskever, and R.~R.
  Salakhutdinov, ``Improving neural networks by preventing co-adaptation of
  feature detectors,'' \emph{arXiv preprint arXiv:1207.0580}, 2012.

\bibitem{wan2013regularization}
L.~Wan, M.~Zeiler, S.~Zhang, Y.~Le~Cun, and R.~Fergus, ``Regularization of
  neural networks using dropconnect,'' in \emph{International Conference on
  Machine Learning}, 2013, pp. 1058--1066.

\bibitem{huang2017snapshot}
G.~Huang, Y.~Li, G.~Pleiss, Z.~Liu, J.~E. Hopcroft, and K.~Q. Weinberger,
  ``Snapshot ensembles: Train 1, get {M} for free,'' in \emph{International
  Conference on Learning Representations}, 2017.

\bibitem{bian2021does}
Y.~Bian and H.~Chen, ``When does diversity help generalization in
  classification ensembles?'' \emph{IEEE Transactions on Cybernetics}, 2021.

\bibitem{he2016deep}
K.~He, X.~Zhang, S.~Ren, and J.~Sun, ``Deep residual learning for image
  recognition,'' in \emph{IEEE Conference on Computer Vision and Pattern
  Recognition}, 2016, pp. 770--778.

\bibitem{huang2017densely}
G.~Huang, Z.~Liu, L.~Van Der~Maaten, and K.~Q. Weinberger, ``Densely connected
  convolutional networks,'' in \emph{IEEE Conference on Computer Vision and
  Pattern Recognition}, 2017, pp. 4700--4708.

\bibitem{liu2021swin}
Z.~Liu, Y.~Lin, Y.~Cao, H.~Hu, Y.~Wei, Z.~Zhang, S.~Lin, and B.~Guo, ``Swin
  {T}ransformer: Hierarchical vision transformer using shifted windows,'' in
  \emph{International Conference on Computer Vision}, 2021, pp.
  10\,012--10\,022.

\bibitem{hou2021vision}
Q.~Hou, Z.~Jiang, L.~Yuan, M.-M. Cheng, S.~Yan, and J.~Feng, ``Vision
  permutator: A permutable {MLP}-like architecture for visual recognition,''
  \emph{IEEE Transactions on Pattern Analysis and Machine Intelligence}, 2022.

\bibitem{freund1996experiments}
Y.~Freund, R.~E. Schapire \emph{et~al.}, ``Experiments with a new boosting
  algorithm,'' in \emph{International Conference on Machine Learning},
  vol.~96.\hskip 1em plus 0.5em minus 0.4em\relax Citeseer, 1996, pp. 148--156.

\bibitem{bian2019ensemble}
Y.~Bian, Y.~Wang, Y.~Yao, and H.~Chen, ``Ensemble pruning based on objection
  maximization with a general distributed framework,'' \emph{IEEE Transactions
  on Neural Networks and Learning Systems}, vol.~31, no.~9, pp. 3766--3774,
  2019.

\bibitem{ren2016ensemble}
Y.~Ren, L.~Zhang, and P.~N. Suganthan, ``Ensemble classification and
  regression-recent developments, applications and future directions,''
  \emph{IEEE Computational Intelligence Magazine}, vol.~11, no.~1, pp. 41--53,
  2016.

\bibitem{ba2013adaptive}
J.~Ba and B.~Frey, ``Adaptive dropout for training deep neural networks,'' in
  \emph{Advances in Neural Information Processing Systems}, 2013, pp.
  3084--3092.

\bibitem{cirecsan2012multi}
D.~Cire{\c{s}}An, U.~Meier, J.~Masci, and J.~Schmidhuber, ``Multi-column deep
  neural network for traffic sign classification,'' \emph{Neural Networks},
  vol.~32, pp. 333--338, 2012.

\bibitem{zhang2016visual}
L.~Zhang and P.~N. Suganthan, ``Visual tracking with convolutional random
  vector functional link network,'' \emph{IEEE Transactions on Cybernetics},
  vol.~47, no.~10, pp. 3243--3253, 2016.

\bibitem{grandvalet2005semi}
Y.~Grandvalet and Y.~Bengio, ``Semi-supervised learning by entropy
  minimization,'' in \emph{Advances in Neural Information Processing Systems},
  2005, pp. 529--536.

\bibitem{lee2016stochastic}
S.~Lee, S.~P.~S. Prakash, M.~Cogswell, V.~Ranjan, D.~Crandall, and D.~Batra,
  ``Stochastic multiple choice learning for training diverse deep ensembles,''
  in \emph{Advances in Neural Information Processing Systems}, 2016, pp.
  2119--2127.

\bibitem{levin1995introduction}
J.~Levin and B.~Nalebuff, ``An introduction to vote-counting schemes,''
  \emph{Journal of Economic Perspectives}, vol.~9, no.~1, pp. 3--26, 1995.

\bibitem{kontschieder2015deep}
P.~Kontschieder, M.~Fiterau, A.~Criminisi, and S.~Rota~Bulo, ``Deep neural
  decision forests,'' in \emph{IEEE International Conference on Computer
  Vision}, 2015, pp. 1467--1475.

\bibitem{rota2014neural}
S.~Rota~Bulo and P.~Kontschieder, ``Neural decision forests for semantic image
  labelling,'' in \emph{IEEE Conference on Computer Vision and Pattern
  Recognition}, 2014, pp. 81--88.

\bibitem{zhou1702deep}
Z.~Zhou and J.~Feng, ``Deep forest: Towards an alternative to deep neural
  networks,'' \emph{arXiv preprint arXiv:1702.08835}, 2017.

\bibitem{pang2018improving}
M.~Pang, K.-M. Ting, P.~Zhao, and Z.-H. Zhou, ``Improving deep forest by
  confidence screening,'' in \emph{IEEE International Conference on Data
  Mining}.\hskip 1em plus 0.5em minus 0.4em\relax IEEE, 2018, pp. 1194--1199.

\bibitem{wenbatchensemble}
Y.~Wen, D.~Tran, and J.~Ba, ``Batch{E}nsemble: Efficient ensemble of deep
  neural networks via rank-1 perturbation,'' in \emph{Advances in Neural
  Information Processing Systems Workshop}, 2019.

\bibitem{shaham2016deep}
U.~Shaham, X.~Cheng, O.~Dror, A.~Jaffe, B.~Nadler, J.~Chang, and Y.~Kluger, ``A
  deep learning approach to unsupervised ensemble learning,'' in
  \emph{International Conference on Machine Learning}, 2016, pp. 30--39.

\bibitem{gu2018regularizing}
S.~Gu, Y.~Hou, L.~Zhang, and Y.~Zhang, ``Regularizing deep neural networks with
  an ensemble-based decorrelation method.'' in \emph{International Joint
  Conference on Artificial Intelligence}, 2018, pp. 2177--2183.

\bibitem{lakshminarayanan2017simple}
B.~Lakshminarayanan, A.~Pritzel, and C.~Blundell, ``Simple and scalable
  predictive uncertainty estimation using deep ensembles,'' in \emph{Advances
  in Neural Information Processing Systems}, 2017, pp. 6402--6413.

\bibitem{gal2016dropout}
Y.~Gal and Z.~Ghahramani, ``Dropout as a bayesian approximation: Representing
  model uncertainty in deep learning,'' in \emph{International Conference on
  Machine Learning}, 2016, pp. 1050--1059.

\bibitem{ritter2018scalable}
H.~Ritter, A.~Botev, and D.~Barber, ``A scalable laplace approximation for
  neural networks,'' in \emph{International Conference on Learning
  Representations}, vol.~6, 2018.

\bibitem{blundell2015weight}
C.~Blundell, J.~Cornebise, K.~Kavukcuoglu, and D.~Wierstra, ``Weight
  uncertainty in neural networks,'' \emph{arXiv preprint arXiv:1505.05424},
  2015.

\bibitem{zhu2018knowledge}
X.~Zhu, S.~Gong \emph{et~al.}, ``Knowledge distillation by on-the-fly native
  ensemble,'' in \emph{Advances in Neural Information Processing Systems},
  2018, pp. 7517--7527.

\bibitem{zheng2018deep}
J.~Zheng, X.~Cao, B.~Zhang, X.~Zhen, and X.~Su, ``Deep ensemble machine for
  video classification,'' \emph{IEEE Transactions on Neural Networks and
  Learning Systems}, vol.~30, no.~2, pp. 553--565, 2018.

\bibitem{chen2021class}
Z.~Chen, J.~Duan, L.~Kang, and G.~Qiu, ``Class-imbalanced deep learning via a
  class-balanced ensemble,'' \emph{IEEE Transactions on Neural Networks and
  Learning Systems}, 2021.

\bibitem{bregman1967relaxation}
L.~M. Bregman, ``The relaxation method of finding the common point of convex
  sets and its application to the solution of problems in convex programming,''
  \emph{USSR Computational Mathematics and Mathematical Physics}, vol.~7,
  no.~3, pp. 200--217, 1967.

\bibitem{banerjee2005clustering}
A.~Banerjee, S.~Merugu, I.~S. Dhillon, and J.~Ghosh, ``Clustering with bregman
  divergences,'' \emph{Journal of Machine Learning Research}, vol.~6, no. Oct,
  pp. 1705--1749, 2005.

\bibitem{simonyan2014very}
K.~Simonyan and A.~Zisserman, ``Very deep convolutional networks for
  large-scale image recognition,'' \emph{arXiv preprint arXiv:1409.1556}, 2014.

\bibitem{szegedy2015going}
C.~Szegedy, W.~Liu, Y.~Jia, P.~Sermanet, S.~Reed, D.~Anguelov, D.~Erhan,
  V.~Vanhoucke, and A.~Rabinovich, ``Going deeper with convolutions,'' in
  \emph{IEEE Conference on Computer Vision and Pattern Recognition}, 2015, pp.
  1--9.

\bibitem{valiant1984theory}
L.~G. Valiant, ``A theory of the learnable,'' \emph{Communications of the ACM},
  vol.~27, no.~11, pp. 1134--1142, 1984.

\bibitem{yu2011diversity}
Y.~Yu, Y.-F. Li, and Z.-H. Zhou, ``Diversity regularized machine,'' in
  \emph{International Joint Conference on Artificial Intelligence}, 2011.

\bibitem{krizhevsky2009learning}
A.~Krizhevsky, G.~Hinton \emph{et~al.}, \emph{Learning multiple layers of
  features from tiny images}.\hskip 1em plus 0.5em minus 0.4em\relax Citeseer,
  2009.

\bibitem{deng2009imagenet}
J.~Deng, W.~Dong, R.~Socher, L.-J. Li, K.~Li, and L.~Fei-Fei, ``Image{N}et: A
  large-scale hierarchical image database,'' in \emph{IEEE Conference on
  Computer Vision and Pattern Recognition}.\hskip 1em plus 0.5em minus
  0.4em\relax Ieee, 2009, pp. 248--255.

\bibitem{paszke2019pytorch}
A.~Paszke, S.~Gross, F.~Massa, A.~Lerer, J.~Bradbury, G.~Chanan, T.~Killeen,
  Z.~Lin, N.~Gimelshein, L.~Antiga \emph{et~al.}, ``Py{T}orch: An imperative
  style, high-performance deep learning library,'' in \emph{Advances in Neural
  Information Processing Systems}, 2019, pp. 8024--8035.

\bibitem{lin2013network}
M.~Lin, Q.~Chen, and S.~Yan, ``Network in network,'' \emph{arXiv preprint
  arXiv:1312.4400}, 2013.

\bibitem{diez2015diversity}
J.~F. D{\'\i}ez-Pastor, J.~J. Rodr{\'\i}guez, C.~I. Garc{\'\i}a-Osorio, and
  L.~I. Kuncheva, ``Diversity techniques improve the performance of the best
  imbalance learning ensembles,'' \emph{Information Sciences}, vol. 325, pp.
  98--117, 2015.

\bibitem{tan2019efficientnet}
M.~Tan and Q.~Le, ``Efficient{N}et: Rethinking model scaling for convolutional
  neural networks,'' in \emph{International Conference on Machine Learning},
  2019, pp. 6105--6114.

\end{thebibliography}
\bibliographystyle{IEEEtran}





\ifCLASSOPTIONcaptionsoff
  \newpage
\fi

\end{document}